\documentclass[journal]{IEEEtran}
\usepackage{amsmath,amsfonts}
\usepackage{algorithmic}
\usepackage{algorithm}
\usepackage{array}
\usepackage[caption=false,font=normalsize,labelfont=sf,textfont=sf]{subfig}
\usepackage{textcomp}
\usepackage{stfloats}
\usepackage{url}
\usepackage{verbatim}
\usepackage{graphicx}
\usepackage[numbers,sort&compress]{natbib}
\usepackage{tabularx,booktabs,multirow}
\usepackage{xcolor}
\usepackage{soul}

\begin{document}

\title{Memory-Augmented Online Test-Time Adaptation for EEG Motor Imagery via Perturbation Consistency and Confidence Calibration}

\author{{Kunkun~Zhang,~Kehao~Wu,~Zhitao~Chen,~Haigen~Hu,~\IEEEmembership{Member,~IEEE},~and~Qianwei~Zhou,~\IEEEmembership{Member,~IEEE}}
\thanks{Received 25 March 2026; revised 21 August 2026; accepted 28 August 2026. Date of publication 22 September 2026; date of current version 5 October 2026. This work was supported in part by the National Natural Science Foundation of China under Grant 62271448 and Grant 62373324 and in part by the ``Pioneer'' and ``Leading Goose'' Research and Development Program of Zhejiang under Grant 2026C02A1221 and Grant 2025C02160. The Associate Editor coordinating the review process was Dr. Susanna Spinsante. \textit{(Corresponding author: Qianwei Zhou.)}}
\thanks{The authors are with the College of Computer Science and Technology, Zhejiang University of Technology, Hangzhou 310023, China, and also with Zhejiang Key Laboratory of Visual Information Intelligent Processing, Hangzhou 310023, China (e-mail: zqw@zjut.edu.cn).}%
\thanks{Digital Object Identifier 10.1109/TIM.2026.3736160}}

\maketitle

\begingroup
\renewcommand\thefootnote{}\footnote{
\noindent © 2026 IEEE. Personal use of this material is permitted. Permission from IEEE must be obtained for all other uses, in any current or future media, including reprinting/republishing this material for advertising or promotional purposes, creating new collective works, for resale or redistribution to servers or lists, or reuse of any copyrighted component of this work in other works. DOI: 10.1109/TIM.2026.3736160
}
\addtocounter{footnote}{-1}
\endgroup

\begin{abstract}
Recently, electroencephalography (EEG)-based brain--computer interfaces (BCIs) have gained traction as portable neural measurement systems. However, the practical deployment of these instruments is often hindered by nonstationarity and cross-subject distribution shifts, which lead to significant intersubject inconsistency. Existing adaptation methods remain impractical for real-time measurement scenarios, as they typically require offline access to source or aggregated target data. To address these challenges, this article proposes memory-augmented online test-time adaptation (OTTA) with perturbation consistency and confidence calibration (MOPC), a source-free framework designed to enhance the reliability of BCI instrumentation. MOPC enables the measurement system to update its decoding parameters on the fly during the inference stage, facilitating single-pass adaptation without accessing sensitive source data. We introduce random-mask perturbations to enforce prediction consistency against inherent EEG signal noise and acquisition artifacts. Furthermore, a confidence-calibrated reweighting loss is designed to integrate feature-space similarity, effectively mitigating unreliable pseudolabels and class imbalance during continuous signal acquisition. To ensure system stability and prevent the catastrophic forgetting of established neural patterns, a lightweight memory module is incorporated to regularize parameter updates within a tight latency budget. Evaluated on the BCI IV IIb and OpenBMI datasets, MOPC achieves state-of-the-art performance, outperforming prior methods by 2.22~\% and 3.67~\%, respectively, while keeping the per-trial adaptation latency to approximately 40--60 ms. This study provides a robust, recalibration-free solution for real-time EEG decoding, marking a significant advancement in the precision and usability of practical BCI measurement instruments. The source code is available at https://github.com/KunKun-Zhang/MOPC.git
\end{abstract}


\section{Introduction}
\IEEEPARstart{B}{rain-computer} interfaces (BCIs) serve as advanced neural measurement instruments, establishing direct pathways between the human brain and external devices for neurorehabilitation and robotics~\cite{Wolpaw2002, NicolasAlonso2012}. Among various paradigms, electroencephalography (EEG)-based motor imagery (MI) is a pivotal measurement technique due to its noninvasiveness and portability~\cite{pfurtscheller2001mi}. By decoding imagined movements from EEG signals, MI-based BCIs can restore motor function in paralyzed patients and enhance human capabilities in virtual environments. These systems have been successfully applied in multiple fields, including wheelchair control~\cite{long2012wheelchair} and lower limb exoskeleton assistance systems~\cite{li2024human}.

Early MI-BCI decoding methods relied on manual feature extraction techniques, such as common spatial patterns (CSPs) and statistical bandpower measures, followed by conventional classifiers~\cite{Ramoser2000, Blankertz2008}. The advent of deep learning has led to the development of end-to-end neural models that automatically learn spatial--temporal representations from raw EEG, resulting in significant improvements in measurement accuracy and robustness~\cite{Schirrmeister2017, lawhern2018eegnet}. Despite these advancements, MI-EEG models trained on a set of subjects (source domain) often experience a sharp decline in decoding reliability when applied to new individuals (target domain)~\cite{kunkun2025ssl}. This is attributed to the differences in signal amplitude and frequency characteristics between subjects~\cite{zhang2021non-invasive-survey}. In metrological terms, the decoder is an instrument whose reliability and reproducibility must hold across subjects and over time, yet this cross-subject variability behaves as instrument drift whose usual remedy, per-subject recalibration, lengthens setup and undermines repeatability~\cite{li2025tale}.

\begin{figure}[!t]
  \centering
  \includegraphics[width=.48\textwidth]{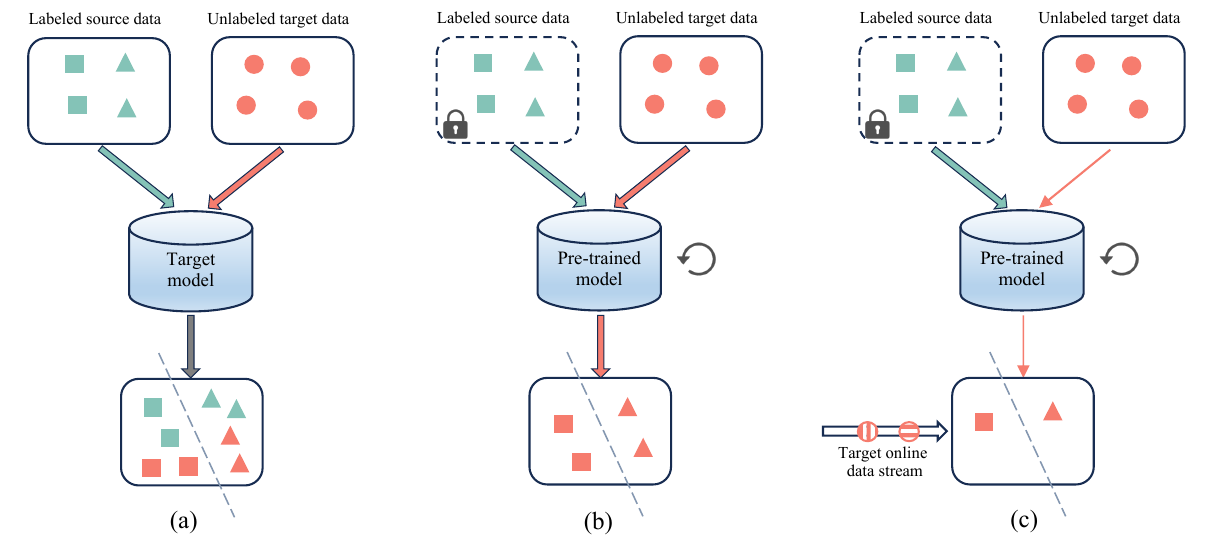}
  \caption{
    Conceptual comparison of three domain-adaptation paradigms for cross-subject EEG decoding.  
    (a) UDA. (b) SFUDA. (c) OTTA.}
  \label{FIG:uda}
\end{figure}

Because such drift degrades any model transferred to a new user, robust transfer learning has become essential~\cite{wu2020transfer-review, zhang2021adaptive-transfer, Mattioli20211dcnn}. Domain adaptation (DA) addresses it by transferring knowledge from a labeled source to an unlabeled target~\cite{farahani2021da-review}, and is commonly divided, depending on whether the source data remain accessible, into unsupervised DA (UDA) and source-free UDA (SFUDA). UDA aligns the two distributions during joint training~\cite{zhang2015deepUDA}. DANN, for instance, learns domain-invariant features with a gradient-reversal adversary~\cite{Ganin2015}. The instrumentation and measurement literature has pursued this line to reconcile cross-subject and cross-device recordings, through multiview contrastive learning~\cite{asgarian2024multiview}, prototype-supervised transfer across multiple source brains (PSATs)~\cite{zhang2024towards}, interdevice sample reweighting (SRENet)~\cite{fu2024improved}, multisource associate alignment across subjects and sessions~\cite{she2023msada}, and semisupervised multisource joint-distribution adaptation that further exploits a few labeled target trials~\cite{jimenez2023ssmjda}, as illustrated in Fig.~\ref{FIG:uda}(a). Instead, SFUDA adapts a pretrained model without revisiting the source, thereby preserving subject privacy~\cite{fang2024sfuda-survey}. SHOT aligns the target with the source hypothesis through information maximization and pseudolabeling~\cite{liang2020shot}, while USFDA adds a generative classifier for out-of-distribution rejection~\cite{kundu2020usfda}, as shown in Fig.~\ref{FIG:uda}(b). Effective as these methods are, they all operate offline: each collects the entire target set, or even a labeled calibration subset of it, and revisits it over several passes before producing a prediction, and the UDA variants further require the source recordings to remain accessible at the deployment site. This assumption does not fit a BCI that must decode a continuous stream one trial at a time.

Online test-time adaptation (OTTA) lifts the offline requirement by updating the model during inference, in a single pass over the incoming data~\cite{liang2025tta-comprehensive, wimpff2024calibration}, which matches the low-latency operation a deployed BCI needs, as depicted in Fig.~\ref{FIG:uda}(c). Existing OTTA methods fall into four families. Entropy minimization sharpens predictions by tuning batch-normalization (BN) parameters, as in TENT~\cite{wang2020tent} and the sharpness-aware SAR~\cite{niu2023sar}. Pseudolabeling supervises the model with its own outputs, as in the paired-view DPLOT~\cite{Yu2024dplot} and the backpropagation-free T3A~\cite{iwasawa2021t3a}. Consistency regularization enforces stable predictions across perturbed views, as used for EEG sleep staging~\cite{guo2025tta-sleep} and for masked self-supervision~\cite{kunkun2025ssl}. Antiforgetting regularization preserves source knowledge over long runs, as in the teacher-based CoTTA~\cite{wang2022cotta} and the probabilistic PETAL~\cite{brahma2023petal}. All four were designed for image streams, however, and overlook two properties of EEG: its trials are short and noisy, and their class balance drifts as the user's engagement varies. The few EEG-specific efforts remain partial. MD-CAT copes with evolving distributions, but by continual retraining rather than test-time adaptation (TTA)~\cite{jin2024affective}; Wang et al.~\cite{wang2024fastrobust} refine a personalized MI decoder trial by trial, retraining on the user's own cue-locked online trials to track day-to-day nonstationarity; Jiao et al.~\cite{jiao2024onlineadaptive} combine subject-specific stimulation and filter-bank settings, optimized offline, with periodic retraining of the decoder during online use; neither considers label-free, source-free transfer to a new user; and T-Time, the closest online baseline, maximizes information over an ensemble yet treats trials as independent, so it cannot exploit their temporal order~\cite{li2023t-time}. Offline prototype networks such as SST-DPN confirm that temporal context is informative~\cite{han2025sst-dpn}, but they cannot adapt once deployed.

\begin{figure*}
  \centering
  \includegraphics[width=0.95\textwidth]{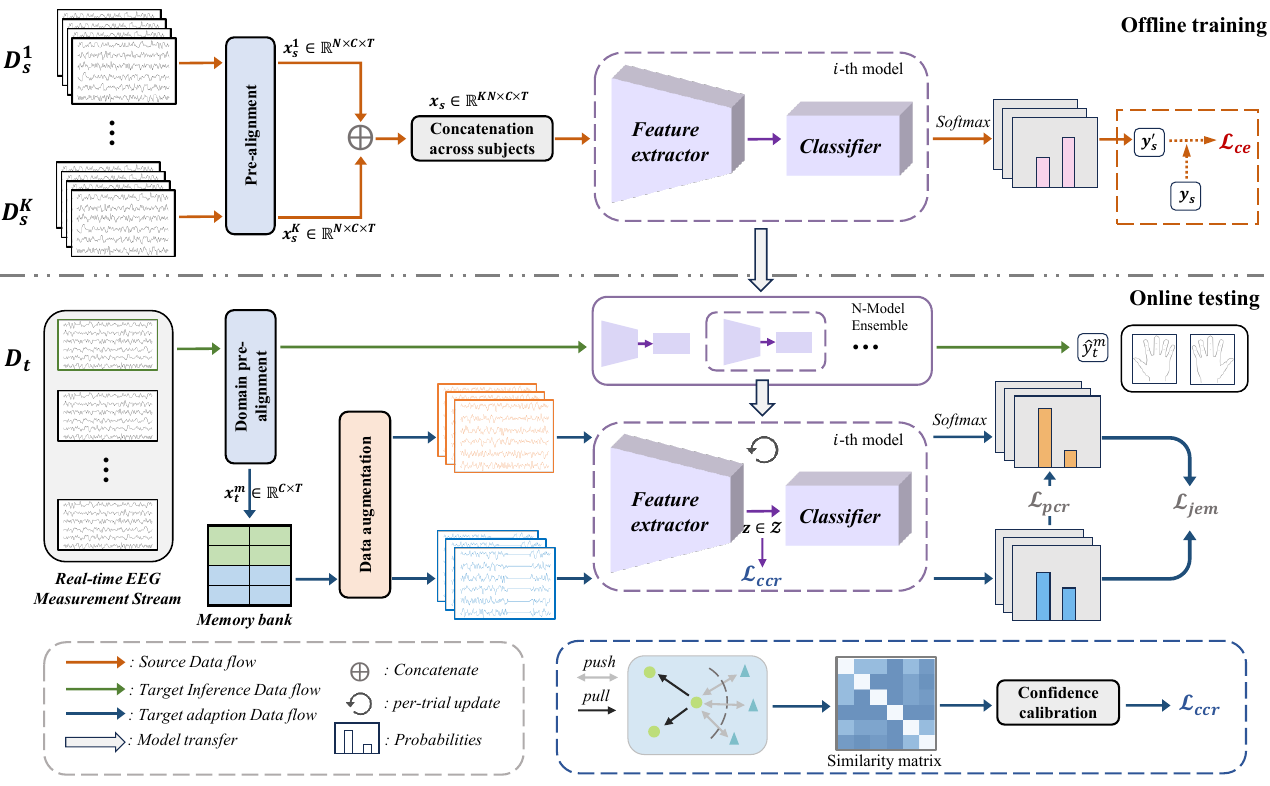}
  \caption{Overview of MOPC. Top (offline training): EEG data from $K$ source subjects are prealigned and concatenated to train an $N$-model ensemble via cross-entropy loss $\mathcal{L}_{\mathrm{ce}}$. Bottom (online testing): For each trial $x_t^{m}$ in the real-time measurement stream: 1) \textit{inference:} the trial is prealigned and evaluated by the ensemble to yield the soft-voted prediction $\hat{y}_t^{m}$, while the aligned data is pushed to the FIFO memory bank; and 2) \textit{adaptation:} a random-mask view is generated to compute $\mathcal{L}_{\mathrm{pcr}}$ and $\mathcal{L}_{\mathrm{jem}}$. Simultaneously, a pairwise feature-similarity matrix is constructed from the memory bank to derive the confidence-calibrated loss $\mathcal{L}_{\mathrm{ccr}}$. The framework minimizes the hybrid loss to update the BN parameters of ensemble members sequentially in a round-robin fashion, ensuring a highly efficient, single-pass adaptation.}
  \label{FIG:framework}
\end{figure*}

We close this gap with MOPC, a memory-augmented OTTA method with perturbation consistency and confidence calibration. MOPC keeps a short first-in-first-out memory of recent prealigned trials together with their predictions, and this memory serves two purposes. First, it continuously re-estimates the Euclidean alignment (EA) of the incoming stream, turning a normally static prealignment~\cite{he2019ea} into an adaptive one. Second, it provides the context for three coupled objectives tailored to EEG: a perturbation-consistency term that keeps the prediction stable when the trial is corrupted by a random time-mask~\cite{kunkun2025ssl}, a joint entropy term that sharpens the predictions on both the original and masked views, and a confidence-calibrated reweighting term that discounts unreliable pseudolabels and offsets class drift using feature similarity among same-class neighbors in memory together with an effective-number class weight~\cite{cui2019class}. Every arriving trial is first labeled by an ensemble of \mbox{$N{=}5$} models and then used to update only the BN parameters of one member, selected in round-robin. Since the update touches nothing else, its cost is independent of the number of classes, which keeps the method fast enough for real-time use. The overall MOPC framework is illustrated in Fig.~\ref{FIG:framework}.

MOPC therefore differs from prior work in both setting and design. Unlike the offline adaptation of PSAT and SRENet or the continual retraining of MD-CAT, it adapts to each trial in a single streaming pass and never revisits the source; unlike the personalized online MI-BCIs of~\cite{wang2024fastrobust} and~\cite{jiao2024onlineadaptive}, which train the decoder on the user's own recordings and refine it by retraining on that user's online trials, it starts from a cross-subject source ensemble and adapts from unlabeled trials in a single pass without any retraining, so it can serve as the decoder-adaptation component of such systems. Unlike the entropy-only, memoryless T-Time, it uses its memory both to stay stable and to follow the changing class balance. Its novelty lies not in any single mechanism, several of which exist in isolation, but in their integration into one lightweight scheme suited to nonstationary EEG. Practically, this targets the outcomes that matter in deployment: it sustains decoding reliability as the signal drifts, reduces the per-subject recalibration burden, and runs on the acquisition hardware in real time, extending the cross-subject transfer line from offline calibration to online operation.

The main contributions of this work are as follows.
\begin{enumerate}
    \item We formulate cross-subject MI decoding as source-free, single-pass OTTA under real-time constraints, in which target trials arrive one at a time and the source is never revisited. This setting is distinct from offline source-free DA and from methods that rely on continual retraining.
    \item We propose MOPC, in which a fixed-capacity FIFO memory drives a dynamic EA of the stream and couples three EEG-aware objectives: a time-masked perturbation-consistency loss, a joint entropy-minimization loss, and a confidence-calibrated reweighting loss that suppresses noisy pseudolabels and corrects class imbalance.
    \item We keep the adaptation lightweight: each trial updates only the BN parameters of a single member of an \mbox{$N{=}5$} ensemble, so the number of trainable parameters does not grow with the number of classes.
\end{enumerate}

The remainder of this article is structured as follows. Section~\ref{sec:method} presents the methodology. Section~\ref{sec:experiments} provides experimental results. Section~\ref{sec:discussion} offers an interpretable analysis of system stability. Section~\ref{sec:conclusion} concludes this article.

\section{Methodology}
\label{sec:method}
\subsection{Preliminaries}
We consider the EEG decoding task as a dynamic measurement problem. Assume that the source domains \( \{ D_s^k \}_{k=1}^{K} \) consist of \( K \) labeled source subjects. For the \( k \)th subject, the source domain \( D_s^k = \{ (x_s^{k,i}, y_s^{k,i}) \}_{i=1}^{n_s^k} \) contains \( n_{s}^{k} \) EEG trials, where \( x_s^{k,i} \in \mathbb{R}^{C \times T} \) represents the \( i \)th EEG signal of the \( k \)th subject, with \( C \) denoting the number of sensor channels and \( T \) the number of time samples, \( y_s^{k,i} \in \{1, \dots, Y\} \) is the corresponding class label. We perform end-to-end training on the pooled data from multiple source subjects to obtain a model \(f_{\theta_0}\). The network comprises a feature extractor \(g_{\phi}\) and a classifier \(h_{\theta}\), i.e., \(f_{\theta_0}=h_{\theta}\circ g_{\phi}\). This means that the \( i \)th EEG trial is first projected into a latent space \( z_i \in \mathbb{R}^{D} \) (where $D$ is the feature dimension) by the feature extractor, and then mapped to a probability vector \( p_i \in \mathbb{R}^{Y} \) in the label space. Concretely, each input trial is a matrix of shape \mbox{$C\times T$}, instantiated as \mbox{$3\times 1000$} on BCI IV IIb and \mbox{$20\times 1000$} on OpenBMI (a 4-s window at 250 Hz); the feature extractor compresses it to a latent vector of dimension \mbox{$D=32$}, and the classifier produces \mbox{$Y$} class scores.

During online adaptation phases, the target domain \( D_t = \{ x_t^{j} \}_{j=1}^{n_t} \) consists of an unlabeled stream of samples from a subject, which is different from the subjects in the source domain. When the \( m \)th EEG signal \( x_t^{m} \) arrives, the proposed OTTA aims to perform an online prediction of \( x_t^{m} \) based on the pretrained model \( f_{\theta_0} \) and the previously received samples \( \{ x_t^{1}, x_t^{2}, \dots, x_t^{m-1} \} \) while updating the model parameters \( \theta_0 \) to adapt to the target domain.

\subsection{Domain Prealignment}
Physiological differences between subjects and device noise lead to nonstationarity in the distribution of EEG trials~\cite{zhang2021non-invasive-survey}. The significant disparity between the source and target domains may result in performance degradation when applying the model directly to the target domain. To reduce this shift, we adopt EA~\cite{he2019ea}, which aligns trial covariances to construct a more stable latent space.

For each source subject \(k\), compute the mean spatial covariance matrix $\overline{R}_{s}^{k} \in \mathbb{R}^{C \times C}$
\begin{equation}
\overline{R}_{s}^{k} = \frac{1}{n_{s}^{k}} \sum_{i=1}^{n_{s}^{k}} x_s^{k,i} (x_s^{k,i})^\top.
\label{Eq: EA}
\end{equation}

Then the inverse square root of the above matrix is taken and applied to each trial in the domain
\begin{equation}
\label{ea_2}
\bar x_{s}^{k,i} = (\overline{R}_{s}^{k})^{-1/2} x_{s}^{k,i}.
\end{equation}

If multiple sessions are available for a subject, EA is performed per session. During online testing, the target domain alignment matrix is estimated from the most recent \(\mathcal{B}\) target trials (see the memory bank below) and updated as the stream evolves.

\subsection{Source Model Training}
Given the constraints of real-time BCI instrumentation, we avoid computationally heavy architectures. We employ EEGNet~\cite{lawhern2018eegnet} as the backbone due to its efficiency in extracting spectral--spatial features. In particular, we configure the first convolutional block with a temporal kernel size of $(1, 64)$ and a depthwise kernel of $(C, 1)$ to capture sensorimotor rhythms. We set the pointwise convolution filter size to 32 and append a global average pooling (GAP) layer. Consequently, the resulting latent feature vector $z_i$ for each trial has a dimension of $\mathbb{R}^{32}$.

To enhance generalization, we apply leave-one-subject-out (LOSO) cross-validation. In each iteration, data from all subjects except the target are used for training with cross-entropy loss $\mathcal{L}_{ce}$. We perform $N=5$-fold cross-validation, retaining the best model from each fold. These $N$ models form an ensemble for robust measurement inference.

\subsection{Online Test-Time Inference and Adaptation}
\subsubsection{Soft Voting Ensemble Learning}
OTTA aims to enable a model to adapt to changes in target domain data after training, particularly when there is a significant discrepancy between the source and target domain sample distributions. To help the model progressively adapt to the target domain data, we adopt a multimodel update strategy. In this approach, model updates occur in rounds, where only one model is updated at a time during each data update. The final prediction is obtained by integrating the softmax probability distributions of multiple models, yielding the most likely class label. In particular, for a sequentially arriving test sample \( x_t^{m} \), the prediction \( \hat{y}_{t}^{m} \) is
\begin{equation}
\label{y_tm}
\hat{y}_{t}^{m} = \arg\max_c \left( \frac{1}{N} \sum_{i=1}^{N} \text{softmax}(f_i(\bar{x}_t^{m}))_c \right)
\end{equation}
where \( N \) is the number of ensemble models and $\bar{x}_t^{m}$ is the prealigned trial (Section~\ref{sec:method}-B). 

\subsubsection{Dynamic Memory Bank}
Streaming EEG exhibits strong temporal dependencies. Updating the model based solely on the current sample makes the model sensitive to transient noise. To address this, we maintain a FIFO memory bank $\mathcal{M}$ with capacity $\mathcal{B}$. Each entry stores the prealigned input \mbox{$\bar{x}_t^{j}$} and its output distribution $p_t^{\,j}$. After the $m$th trial has arrived, the bank is defined as
\begin{equation}
\label{eq:bank}
\mathcal{M} = \{(\bar{x}_t^{j}, p_t^{j})\}_{j = \max(1, m - \mathcal{B} + 1)}^{m}.
\end{equation}
Upon arrival of a new trial, the oldest entry is discarded to keep $|\mathcal{M}|=\mathcal{B}$. The memory capacity is set to $\mathcal{B}=24$, which empirically balances the tradeoff between tracking recent distribution drifts and maintaining sufficient statistics for stable updates (see Fig. \ref{FIG:bank}).

\subsubsection{Perturbation Consistency Regularization}
To improve the model's robustness during the OTTA and prevent overfitting to noise and perturbations in the target domain, we introduce perturbation consistency regularization. This regularization term enhances the model's adaptation to target domain data by enforcing consistency in predictions between the original EEG signal and its augmented version in the output space.

We employ a random masking data augmentation strategy to simulate information loss and noise perturbations in the signal. In particular, a random time point \( \tau \) is initialized from the time interval \( [0, T-W] \), and a local EEG signal with a masking window of length \( W \) is used for augmentation, such that \( \bar{x}_t^{j}[1 : C, \tau : \tau + W] = 0 \). The augmented sample is then denoted as \( \tilde{x}_t^{j} \). In all main experiments, we use \( W=200 \) as the default time samples.

For the original signal \( \bar{x}_t^{j} \) and its augmented version \( \tilde{x}_t^{j} \) in the memory bank, we define the perturbation consistency regularization by calculating the Kullback--Leibler (KL) divergence between their output probability distributions
\begin{equation}
\label{l_pcr}
\mathcal{L}_{pcr} = \frac{1}{\mathcal{B}} \sum_{j=1}^{\mathcal{B}} D_{KL}(f_\theta(\bar{x}_t^{j}) || f_\theta(\tilde{x}_t^{j})).
\end{equation}

In our experiments, compared to cross-entropy loss~\cite{wang2022cotta, Yu2024dplot}, the KL divergence as a consistency loss yields better performance.

\subsubsection{Joint Entropy Minimization}
Entropy minimization is widely used in TTA methods such as TENT~\cite{wang2020tent} to reduce the prediction uncertainty on target domain samples, thus improving the model's confidence. This process helps improve intraclass feature consistency and narrows the domain gap between the source and target domains, enhancing the model's generalization ability on the target domain.

In our approach, we combine the original data and its augmented version (generated via perturbation) in the entropy minimization process. The joint loss is calculated by minimizing the Shannon entropy of both the original signal \( \bar{x}_t^{j} \) and its augmented version \( \tilde{x}_t^{j} \), as follows:
\begin{equation}
\label{l_jem}
\mathcal{L}_{\text{jem}} = - \frac{1}{\mathcal{B}} \sum_{j=1}^{\mathcal{B}} \sum_{c} \left[ \hat{y}_{c}^{j} \log \hat{y}_{c}^{j} + \hat{y}_{c}^{\tilde{j}} \log \hat{y}_{c}^{\tilde{j}} \right]
\end{equation}
where \( \hat{y}_{c}^{j} \) and \( \hat{y}_{c}^{\tilde{j}} \) represent the predicted probabilities for class \( c \) on \( \bar{x}_t^{j} \) and \( \tilde{x}_t^{j} \), respectively.

\subsubsection{Confidence-Calibrated Reweighting}
In online EEG MI decoding, the class proportions in the target stream may drift away from the training prior, and pseudolabel noise can distort the predicted class distribution~\cite{niu2023sar}. We address both issues with a confidence-calibrated reweighting loss \( \mathcal{L}_{ccr} \) that emphasizes reliable and minority samples.

We first derive a confidence weight per sample by comparing latent features pairwise based on the historical data stored in the memory bank, and aggregating similarities from same-class neighbors. The latent features themselves are recalculated each time dynamically, based on the current sample and the stored historical data. Let \( z_j \) be the latent feature of the trial \( j \) and \( S_{ji}=\cos(z_j,z_i) \) the cosine similarity. The confidence weight is
\begin{equation}
\label{c_j}
\hat{C}_j = \frac{\sum_{i=1}^{\mathcal{B}} \exp(S_{ji}) \times M_{ji}}{\sum_{i=1}^{\mathcal{B}} M_{ji} + \epsilon}
\end{equation}
where \( \epsilon \) is a small constant to prevent division by zero and \( M_{ji} \) is the class similarity mask
\begin{equation}
M_{ji} = \begin{cases}
1, & \text{if } y_j = y_i \text{ and } j \neq i \\
0, & \text{else.}
\end{cases}
\end{equation}

To mitigate class imbalance under nonstationary priors, we adopt the effective-number weighting~\cite{cui2019class}
\begin{equation}
\label{a_t}
\alpha_{\hat{y}_j} = \frac{1 - \beta}{1 - \beta^{n_{\hat{y}_j}}}
\end{equation}
where \( n_{\hat{y}_j} \) is the running pseudocount of class \( \hat{y}_j \) observed in the stream so far and \( \beta\in[0,1) \) controls the degree of balancing.
Combining confidence, class balance, and focal modulation yields
\begin{equation}
\label{l_ccr}
\mathcal{L}_{ccr} = - \frac{1}{\mathcal{B}} \sum_{j=1}^{\mathcal{B}} \left[\hat{C}_j \cdot \alpha_{\hat{y}_j} (1 - p(\hat{y}_{j}))^\gamma \log(p(\hat{y}_{j})) \right]
\end{equation}
where \( p(\hat{y}_j) \) is the predicted probability of the pseudolabel for trial \( j \) and \( \gamma>0 \) is the focal parameter. This formulation downweights unreliable pseudolabels, upweights informative minority and hard samples, and stabilizes adaptation under class-imbalanced, nonstationary test streams.

\subsection{Optimization of MOPC}
To jointly optimize MOPC in the streaming setting, we minimize a hybrid objective that combines perturbation consistency regularization, joint entropy minimization, and confidence-calibrated reweighting
\begin{equation}
\label{l_total}
\mathcal{L}_{\text{total}}=\mathcal{L}_{\text{pcr}}+\mathcal{L}_{\text{jem}}+\mathcal{L}_{\text{ccr}}.
\end{equation}
For each incoming trial, we first obtain its prediction and subsequently perform a single adaptation step via one gradient update with $\mathcal{L}_{\text{total}}$ while keeping all network weights fixed except for the BN layers. The update touches only the affine scale-and-shift parameters of the BN layers in the selected ensemble member, while the convolutional kernels, the depthwise and pointwise filters, and the classifier weights remain frozen. Each trial is thus handled in one streaming pass: one soft-voted ensemble prediction over the \mbox{$N=5$} members, followed, once the memory bank is populated, by a single BN-only gradient step over the \mbox{$\mathcal{B}=24$} memory entries, with no trial ever revisited. This per-trial schedule couples inference with adaptation to enable continuous online learning. Algorithm \ref{alg:mopc} summarizes this schedule.

\begin{algorithm}[!t]
\caption{Proposed MOPC Framework}
\label{alg:mopc}
\begin{algorithmic}
\renewcommand{\algorithmicrequire}{\textbf{Input:}}
\renewcommand{\algorithmicensure}{\textbf{Output:}}
\REQUIRE 
Labeled source domain data $\{D_s^k\}_{k=1}^K$;\\
Initial $N$-model ensemble $\{f_{\theta_i}\}_{i=1}^N$;\\
Streaming target data $\{x_t^m\}_{m=1}^{n_t}$;\\
$\mathcal{B}$, the memory bank capacity;\\
\ENSURE 
The classification $\hat{y}_t^m$ for $x_t^m$.

\STATE // \textit{Stage 1: Offline Training}
\FOR{$i = 1 : N$}
    \STATE Train model $f_{\theta_i}$ on $N-1$ folds using $\mathcal{L}_{ce}$;
    \STATE Retain the best-performing model to build the ensemble $\{f_{\theta_i}\}_{i=1}^N$;
\ENDFOR

\STATE // \textit{Stage 2: Online Testing}
\STATE Initialize the dynamic memory bank $\mathcal{M} = \emptyset$;
\FOR{$m = 1 : n_t$}
    \STATE // \textit{Online Inference}
    \IF{$|\mathcal{M}| < \mathcal{B}$}
        \STATE Align $x_t^m$ via source EA matrix to get $\bar{x}_t^m$;
    \ELSE
        \STATE Calculate the covariance matrix on $\mathcal{M}$ by \eqref{Eq: EA};
        \STATE Perform EA on $x_t^m$ to get $\bar{x}_t^m$;
    \ENDIF
    \STATE Calculate $p_t^m$ and predicted label $\hat{y}_t^m$ by \eqref{y_tm};
    \STATE Append $(\bar{x}_t^m, p_t^m)$ to $\mathcal{M}$;
    \IF{$|\mathcal{M}| > \mathcal{B}$}
        \STATE Dequeue the oldest trial from $\mathcal{M}$;
    \ENDIF

    \STATE // \textit{Online Adaptation}
    \IF{$|\mathcal{M}| == \mathcal{B}$}
        \STATE Select the ensemble member $i = (m \bmod N) + 1$;
        \STATE Calculate $\mathcal{L}_{pcr}$ on $\mathcal{M}$ by \eqref{l_pcr};
        \STATE Calculate $\mathcal{L}_{jem}$ on $\mathcal{M}$ by \eqref{l_jem};
        \STATE Calculate $\mathcal{L}_{ccr}$ on $\mathcal{M}$ by \eqref{l_ccr};
        \STATE Calculate total loss $\mathcal{L}_{\text{total}}$ by \eqref{l_total} and update the BN parameters $\theta_{i}^{BN}$ of $f_{\theta_i}$ using $\nabla \mathcal{L}_{\text{total}}$;
    \ENDIF
\ENDFOR
\end{algorithmic}
\end{algorithm}

\section{Experiments and Results}
\label{sec:experiments}
\subsection{Datasets}
To evaluate the classification performance of our framework in the OTTA setting, we conducted experiments using two public datasets: BCI Competition IV IIb and OpenBMI. Because each benchmark defines its own measurement front-end, we detail below the acquisition instrumentation, electrode montage, and sampling resolution of the two datasets, complementing the processing pipeline summarized in Table~\ref{tab:config}. A third dataset, the four-class BCI Competition IV 2a, is introduced only as the basis for the preliminary multiclass validation of Section~\ref{subsec:multiclass}, and is not part of the binary experiments.

\subsubsection{BCI IV IIb}
The BCI IV IIb~\cite{leeb2008bci2b} dataset contains EEG signals from 9 healthy subjects (B01--B09), recorded across five separate sessions at a sampling rate of 250 Hz using 3 EEG channels (around C3/Cz/C4 under the 10–20 system). Each subject performed two MI tasks: left-hand (LH) and right-hand (RH) movement imagery. The first two sessions consisted of 120 trials each, without feedback, while the last three sessions included 160 trials each, with a smiley face displayed on the screen as feedback. In our experiments, we extract a 4-s segment of EEG data following the visual cue as an EEG trial. At acquisition, these three derivations were recorded as bipolar channels, accompanied by three monopolar EOG channels retained solely for artifact processing and never used for decoding; Fz served as the ground. The amplifier applied a 0.5--100-Hz recording bandpass and a 50-Hz notch, with its dynamic range set to \mbox{$\pm100\,\mu\mathrm{V}$} for the two no-feedback sessions and \mbox{$\pm50\,\mu\mathrm{V}$} for the three feedback sessions. Each 4-s cue-locked trial therefore corresponds to a \mbox{$1000$}-sample analysis window, evenly split between the left- and RH classes, and the cross-subject protocol yields about 5760 pooled source trials with 320 test trials per subject.

\subsubsection{OpenBMI}
The OpenBMI~\cite{lee2019openbmi} dataset contains multisession, multiparadigm EEG signals recorded from 54 healthy subjects (subj01-subj54) at a sampling rate of 1000 Hz, using 62 EEG electrode channels. In this study, we focused on the MI paradigm using a subset of 20 channels located over the FC, C, CP, and P rows to balance spatial resolution and computational latency~\cite{lee2019openbmi}. Each subject participated in two sessions, each consisting of an offline training phase and an online testing phase, with 100 trials in each phase. We consider the EEG data from [0, 4] seconds after the visual cue as a trial. Acquisition used a BrainAmp amplifier (Brain Products, Munich, Germany) with Ag/AgCl electrodes placed according to the international 10--20 system. Recordings were nasion-referenced and grounded at AFz, with electrode impedance kept below \mbox{$10\,\mathrm{k}\Omega$}. After the down-sampling summarized in Table~\ref{tab:config}, the 0--4-s cue-locked segment gives a \mbox{$1000$}-sample analysis window, evenly split between the two motor-imagery classes, for a pooled source set of 21\,200 trials and 200 test trials per subject.

\subsubsection{BCI IV 2a}
The four-class BCI Competition IV 2a dataset~\cite{brunner2008bci} is the multiclass sibling of BCI IV IIb, recorded at the same 250-Hz sampling rate under the shared Graz acquisition protocol. It comprises 9 subjects (A01--A09) and, unlike the three bipolar derivations of IIb, provides 22 EEG channels and four motor-imagery classes (left hand, right hand, feet, and tongue), with 288 trials in each of two sessions (training and evaluation). As in the other datasets, we take the 0--4 s cue-locked segment, yielding a \mbox{$1000$}-sample analysis window at 250 Hz. This dataset is used solely for the preliminary multiclass validation of Section~\ref{subsec:multiclass}, and not for the main binary experiments.

\subsection{Data Preprocessing}
Before feeding the raw digitized signals into the MOPC framework, a rigorous preprocessing pipeline was implemented to minimize measurement noise and cross-session instabilities. For the BCI IV IIb dataset, we further apply a fifth-order Butterworth bandpass filter with a passband of 4--40 Hz for analysis. For the OpenBMI dataset, we apply a bandpass filter in the range of 8--30 Hz~\cite{lee2019openbmi}, and downsample the data to 250 Hz to reduce the data size and improve computational efficiency during online processing~\cite{wang2023ifnet}. To remove amplitude discrepancies between trials, filtered EEG signals are normalized using the following trial normalization before applying Euclidean calibration~\cite{miao2023lmda}:
\begin{equation}
x_i = \frac{x_i}{\max(|x_i|)}.
\end{equation}

\subsection{Experiment Settings}
To evaluate the generalization capability across different individuals, we employ LOSO cross-validation to simulate cross-subject deployment. In each iteration, one subject’s data is treated as the unlabeled target domain (test set), while the labeled data from all other subjects form the source domain. The labeled source domain is then divided into \(N=5\) folds for cross-validation. In each of the \(N\) rounds, one fold is held out for validation, while the remaining \(N-1\) folds are used for training, resulting in \(N\) pretrained source-domain models. The hyperparameters for the \(\mathcal{L}_{ccr}\) were set to \mbox{$\beta = 0.999$} following the effective-number weighting of~\cite{cui2019class}, and \mbox{$\gamma = 0.5$} as a standard focal-modulation setting.

All sessions from the source subjects are pooled for training and validation, and only the test sessions from the target subject are used for testing. Under this setup, the BCI IV IIb source set contains approximately 5760 trials, and the test set contains 320 trials per subject. For OpenBMI, the source set comprises 21\,200 trials, and the test set contains 200 trials per subject.

The source domain models are trained using the PyTorch framework on an Intel\footnote{Registered trademark.} Xeon\footnotemark[\value{footnote}] Silver 4210R CPU @ 2.40~GHz and an NVIDIA RTX A6000 GPU. Offline, all network parameters of each EEGNet instance are trained with $\mathcal{L}_{ce}$ by AdamW under the cosine-annealing schedule of Table~\ref{tab:config}, and the checkpoint with the best accuracy on the held-out inner fold is retained; neither training nor checkpoint selection ever sees the target subject, and the procedure is repeated over five random seeds.

In the target DA process, simulating an online scenario, all updates run on the CPU and use the same 4-s trial window as offline training (\mbox{$T=1000$} samples at 250 Hz). Adaptation is deliberately conservative. Per trial, a single AdamW step with a reduced learning rate of \(6 \times 10^{-4}\) updates only the BN affine parameters of one ensemble member, selected in round-robin, with the loss $\mathcal{L}_{\text{total}}$ evaluated over the $\mathcal{B}$ entries of the memory bank. Each ensemble member keeps its own optimizer; no optimizer state is carried over from the offline stage, and every trial is predicted before the update it triggers and is never revisited. The hyperparameters required no tuning beyond $\mathcal{B}$: $\beta$ and $\gamma$ follow the settings given above, the mask length was fixed a priori to $W=200$ samples ($0.8$ s, one fifth of the trial window), and the three losses are summed with equal weights. The capacity $\mathcal{B}$ itself was chosen from the sweep of Fig.~\ref{FIG:bank} as a single value shared by both datasets and all subjects. No hyperparameter is tuned per subject or per dataset, and the same configuration is reused for the class-imbalanced protocol, the other backbones, and the four-class experiment. Table~\ref{tab:config} consolidates the complete configuration so that the pipeline can be replicated from a single reference.

\begin{table*}[!t]
  \centering
  \caption{Consolidated experimental configuration for reproducibility. All values are those reported in the text; the table gathers them into a single reference}
  \label{tab:config}
  \small
  \renewcommand{\arraystretch}{1.15}
  \begin{tabular}{@{}l >{\raggedright\arraybackslash}p{0.30\textwidth} >{\raggedright\arraybackslash}p{0.38\textwidth}@{}}
    \hline\hline
    Parameter & Value & Note \\
    \hline
    \multicolumn{3}{@{}l}{\textit{Signal acquisition and preprocessing}}\\
    Sampling rate & 250 Hz (IIb); 1000$\rightarrow$250 Hz (OpenBMI) & OpenBMI down-sampled for online efficiency \\
    EEG channels $C$ & 3 (IIb: C3/Cz/C4); 20 (OpenBMI) & OpenBMI uses a 20-channel subset of 62 \\
    Trial window & 4 s $= 1000$ samples ($T$) & $0$--$4$ s after the cue, at 250 Hz \\
    Band-pass filter & 4--40 Hz, 5th-order Butterworth (IIb); 8--30 Hz (OpenBMI) &  \\
    \hline
    \multicolumn{3}{@{}l}{\textit{Offline source-model training}}\\
    Optimizer & AdamW & lr $1\times10^{-3}$ (cosine annealing), weight decay $1\times10^{-5}$ \\
    Epochs / batch size & 1000 / 32 &  \\
    Cross-validation & LOSO (outer), 5-fold (inner) & one model per fold; ensemble size $N=5$ \\
    Random seeds & 5 & cuDNN deterministic mode \\
    \hline
    \multicolumn{3}{@{}l}{\textit{Online test-time adaptation}}\\
    Updated parameters & BN affine (scale/shift) only & all other weights frozen; $\mathcal{O}(1)$ in class count \\
    Adaptation optimizer & AdamW, lr $6\times10^{-4}$ & one gradient step per trial, on CPU \\
    Memory bank size $\mathcal{B}$ & 24 & FIFO \\
    Mask window $W$ & 200 samples ($0.8$ s) & random time mask for consistency \\
    Class-balance / focal ($\beta,\gamma$) & 0.999 / 0.5 & effective-number weighting; focal modulation \\
    \hline\hline
  \end{tabular}
\end{table*}

\subsection{Baseline Methods}
To rigorously assess the measurement reliability of the MOPC framework, we benchmarked it against three distinct categories of transfer learning methods: direct transfer (DT), DA, and TTA. The DT baselines include three end-to-end models: EEGNet~\cite{lawhern2018eegnet}, ConvNet~\cite{Schirrmeister2017}, and LMDANet~\cite{miao2023lmda}. The DA group consists of the unsupervised domain adaptation (UDA) method DANN~\cite{Ganin2015} and the source-free unsupervised domain adaptation (SFUDA) method SHOT~\cite{liang2020shot}. The TTA baselines include four established methods: TENT~\cite{wang2020tent}, CoTTA~\cite{wang2022cotta}, T-Time~\cite{li2023t-time}, and DPLOT~\cite{Yu2024dplot}. All methods are evaluated under the same LOSO protocol with balanced classes, and we report cross-subject performance metrics, including accuracy, F1-score, Cohen’s kappa, and standard deviation. For all transfer learning baseline methods, EEGNet is used as the backbone, and the same domain prealignment (DPA) procedure is applied during model transfer, so that the adaptation mechanism is the only factor that differs between the DA and TTA rows of Tables~\ref{tab:bci2b}--\ref{tab:imbalanced_class_accuracy}; the absolute values are therefore not meant to be compared with those reported in the original papers, which use other datasets, splits, or backbones.

\begin{table*}[!t]
  \centering
  \caption{Class-balanced cross-subject classification accuracies on BCI IV IIb (\%)}
  \label{tab:bci2b}
  \small
  \renewcommand{\arraystretch}{1.20}
  \setlength{\tabcolsep}{4.2pt}
  \begin{tabular*}{\textwidth}{@{\extracolsep{\fill}} c l c c c c c c c c c c @{}}
    \hline\hline
    Category & Method & B01 & B02 & B03 & B04 & B05 & B06 & B07 & B08 & B09 & Avg. acc $\pm$ std \\
    \hline
    \multirow{3}{*}{DT}
      & EEGNet  & 73.12 & 61.79 & 56.87 & 88.75 & 80.94 & 75.94 & 76.88 & 84.38 & \underline{87.81} & 76.28 $\pm$ 11.03\textsuperscript{**} \\
      & ConvNet & 63.75 & 59.29 & 54.69 & 93.75 & 71.25 & 74.38 & 74.06 & 80.63 & 86.25 & 73.12 $\pm$ 12.64\textsuperscript{**} \\
      & LMDANet & 72.81 & 60.00 & 55.94 & 90.31 & 73.12 & 82.50 & 76.56 & 81.87 & 86.56 & 75.52 $\pm$ 11.55\textsuperscript{**} \\
    \hline
    \multirow{2}{*}{DA}
      & DANN & 73.75 & 62.50 & 57.50 & 89.69 & \textbf{86.25} & 76.88 & 76.56 & 84.06 & 87.50 & 77.19 $\pm$ 11.22\textsuperscript{**} \\
      & SHOT & \underline{75.31} & 59.29 & \underline{57.81} & 95.94 & 84.38 & 74.38 & \underline{80.94} & 90.62 & 86.25 & 78.32 $\pm$ 13.11\textsuperscript{*} \\
    \hline
    \multirow{5}{*}{TTA}
      & TENT   & 73.12 & 57.86 & 54.69 & \underline{96.56} & 83.44 & 78.75 & 78.75 & 89.69 & 84.38 & 77.47 $\pm$ 13.80\textsuperscript{**} \\
      & CoTTA  & \underline{75.31} & \underline{63.57} & 55.31 & 93.75 & 81.25 & 78.12 & 78.75 & \underline{91.56} & 86.88 & 78.28 $\pm$ 12.51\textsuperscript{*} \\
      & T-Time & \underline{75.31} & 58.57 & \textbf{59.06} & 95.31 & 84.38 & \underline{82.50} & 79.37 & 89.38 & 87.19 & \underline{79.01 $\pm$ 12.80}\textsuperscript{*} \\
      & DPLOT  & 75.00 & 63.21 & 56.25 & 94.37 & 79.06 & 73.75 & \underline{80.94} & 90.94 & 85.62 & 77.68 $\pm$ 12.37\textsuperscript{**} \\
      & MOPC   & \textbf{79.37} & \textbf{63.57} & \underline{57.81} & \textbf{96.88} & \underline{85.94} & \textbf{84.69} & \textbf{81.25} & \textbf{91.87} & \textbf{89.69} & \textbf{81.23 $\pm$ 12.88} \\
    \hline\hline
  \end{tabular*}
  \begin{flushleft}
    \footnotesize The best accuracies are marked in bold, and the second best by an underline.\\
    \textsuperscript{*}: $p<0.05$, \textsuperscript{**}: $p<0.01$, \textsuperscript{***}: $p<0.001$; Wilcoxon signed-rank test comparing each baseline to ours.
  \end{flushleft}
\end{table*}

\begin{table}[!t]
  \centering
  \caption{Class-balanced cross-subject classification accuracies on OpenBMI (\%)}
  \label{tab:openbmi}
  \small
  \renewcommand{\arraystretch}{1.12}
  \setlength{\tabcolsep}{6pt}
  \begin{tabular}{c l c c c c}
    \hline\hline
    Category & Method & Average & Std & F1 & Kappa \\
    \hline
    \multirow{3}{*}{DT}
      & EEGNet  & 73.57\textsuperscript{***} & 12.66 & 74.71 & 47.14 \\
      & ConvNet & 72.83\textsuperscript{***} & 12.92 & 73.85 & 45.66 \\
      & LMDANet & 70.17\textsuperscript{***} & 15.52 & 68.76 & 40.33 \\
    \hline
    \multirow{2}{*}{DA}
      & DANN & 73.82\textsuperscript{***} & 12.71 & 74.76 & 47.64 \\
      & SHOT & 74.32\textsuperscript{**}  & 12.53 & 75.14 & 48.64 \\
    \hline
    \multirow{5}{*}{TTA}
      & TENT   & 74.21\textsuperscript{***} & \underline{12.49} & 75.31 & 48.42 \\
      & CoTTA  & 74.07\textsuperscript{***} & 12.52 & 74.21 & 48.14 \\
      & T-Time & \underline{75.27}\textsuperscript{**} & 12.60 & \underline{75.37} & \underline{50.54} \\
      & DPLOT  & 75.05\textsuperscript{***} & 12.71 & 74.10 & 50.10 \\
      & MOPC   & \textbf{78.94} & \textbf{11.59} & \textbf{78.39} & \textbf{57.88} \\
    \hline\hline
  \end{tabular}
  \begin{flushleft}
    \footnotesize The best accuracies are marked in bold, and the second best by an underline.\\
    \textsuperscript{*}: $p<0.05$, \textsuperscript{**}: $p<0.01$, \textsuperscript{***}: $p<0.001$; Wilcoxon signed-rank test comparing each baseline to ours.
  \end{flushleft}
\end{table}

\subsection{Results}
\subsubsection{Class-Balanced Accuracy Evaluation}
MOPC consistently outperforms DT, DA, and TTA baselines on both the BCI IV IIb and OpenBMI datasets, as reported in Tables~\ref{tab:bci2b} and~\ref{tab:openbmi}. On BCI IV IIb, MOPC achieves an average accuracy of 81.23~\%, exceeding the best direct-transfer model EEGNet by 4.95~\%, surpassing the leading domain-adaptation method SHOT by 2.91~\%, and improving on the strongest TTA baseline T-Time by 2.22~\%. These gains are observed for nearly all subjects, indicating robust cross-subject generalization and stable performance. On OpenBMI, MOPC attains an average accuracy of 78.94~\%, which represents a 3.67~\% improvement over T-Time at 75.27~\%. In addition, MOPC reduces cross-subject variability and increases the F1-score to 78.39~\%, while Cohen’s kappa rises to 57.88, reflecting more reliable and balanced decoding across users. Wilcoxon signed-rank tests~\cite{wilcoxon1992p-value} on paired accuracies confirm that all improvements are significant (p-value $<$ 0.05).

The per-subject columns of Table~\ref{tab:bci2b} show where these gains land: MOPC wins 77 of the 81 paired comparisons against the nine baselines, with two ties and two losses of at most 1.25 points. Its margins are smallest on B02 and B03, whose source-only accuracies are near chance, and largest (7.5--8.8 points over EEGNet) on B04, B06, and B08, whose source models already partially separate the classes. This is the expected behavior of a label-free method: the three objectives can realign a decision boundary that the source representation encodes, as Fig.~\ref{fig:tsne} illustrates for B04 and B06, but cannot create class structure that it lacks. The one subject where T-Time leads, B03, lies in this near-chance regime.

The mechanisms behind the family level ordering are also visible. Although SHOT revisits the entire target session over several epochs, it still trails the single-pass MOPC by 2.9 and 4.6 points on the two datasets. Thus, for nonstationary EEG, the most recent $\mathcal{B}$ trials are more informative than global target statistics. TENT updates the same BN affine parameters as MOPC but uses only the current trial, and it gains far less over EEGNet (1.2 and 0.6 points, against 5.0 and 5.4 for MOPC). Removing the memory bank from MOPC reproduces this drop (see Section~\ref{subsec:ablation}), identifying the memory-driven context as the primary source of the gain. Against the strongest baseline, T-Time, which treats trials independently, MOPC further raises kappa on OpenBMI by 7.3 points. It also attains the lowest cross-subject standard deviation of all ten methods (11.59) while raising the mean, so the improvement is not carried by a few well-performing subjects.

\begin{table}[!t]
  \centering
  \caption{Class-imbalanced cross-subject accuracy (\%) on BCI IV IIb and OpenBMI}
  \label{tab:imbalanced_class_accuracy}
  \small
  \renewcommand{\arraystretch}{1.12}        
  \setlength{\tabcolsep}{8pt}               
  \begin{tabular}{c l c c}
    \hline\hline
    Category & Method & BCI IV IIb & OpenBMI \\
    \hline
    \multirow{3}{*}{DT}
      & EEGNet  & 76.75 $\pm$ 0.76 & 74.99 $\pm$ 0.43 \\
      & ConvNet & 74.96 $\pm$ 0.83 & 73.42 $\pm$ 0.57 \\
      & LMDANet & 76.31 $\pm$ 0.63 & 72.67 $\pm$ 0.39 \\
    \hline
    \multirow{2}{*}{DA}
      & DANN & 76.20 $\pm$ 0.72 & 74.31 $\pm$ 0.53 \\
      & SHOT & 77.21 $\pm$ 0.57 & 75.21 $\pm$ 0.37 \\
    \hline
    \multirow{5}{*}{TTA}
      & TENT   & 76.93 $\pm$ 0.64 & 75.17 $\pm$ 0.46 \\
      & CoTTA  & 77.53 $\pm$ 0.81 & 75.29 $\pm$ 0.34 \\
      & T-Time & \underline{77.77 $\pm$ 0.44} & \underline{75.73 $\pm$ 0.76} \\
      & DPLOT  & 77.41 $\pm$ 0.67 & 75.34 $\pm$ 0.55 \\
      & MOPC   & \textbf{78.86 $\pm$ 0.46} & \textbf{78.78 $\pm$ 0.45} \\
    \hline\hline
  \end{tabular}
  \begin{flushleft}
    \footnotesize The best accuracies are marked in bold, and the second best by an underline.
  \end{flushleft}
\end{table}

\subsubsection{Class-Imbalanced Accuracy Evaluation}
To emulate a class imbalance scenario during test time, we randomly removed half of the samples from one class in each target test set, creating a 2:1 ratio between the two motor-imagery classes. In particular, for BCI IV IIb we removed 50~\% of RH trials, and for OpenBMI we removed 50~\% of LH trials. The adaptation protocol and hyperparameters were identical to those used in the balanced setting. As reported in Table~\ref{tab:imbalanced_class_accuracy}, MOPC consistently surpasses all baselines under this imbalance, outperforming T-Time (which includes a marginal-distribution regularizer intended for imbalanced test data) by 1.09~\% and 3.05~\% on BCI IV IIb and OpenBMI datasets, respectively. 

Because plain accuracy on an imbalanced set is dominated by the majority class, Table~\ref{tab:imbalanced_class_accuracy} should be read together with the balanced results: the ranking of the two strongest methods and of the direct-transfer models is preserved, and MOPC's margin over the strongest baseline stays close to its balanced value on OpenBMI (3.05 versus 3.67 points) while narrowing on BCI IV IIb (1.09 versus 2.22 points), so the adaptation gain is retained rather than traded when the class prior drifts. Consistently, removing $\mathcal{L}_{ccr}$ causes the largest single-term drop in Table~\ref{tab:ablation_results} on both datasets ($-1.65$ and $-1.78$ points), identifying the confidence-calibrated reweighting as the component that counters the drifting prior.

\subsection{Ablation Experiments}
\label{subsec:ablation}
\subsubsection{Component Ablation Study}
To quantify the contributions of each loss component, we ablated \( \mathcal{L}_{pcr} \), \( \mathcal{L}_{jem} \), and \( \mathcal{L}_{ccr} \) in turn on BCI IV IIb. In Table~\ref{tab:ablation_results}, baseline denotes the nonadaptation setting, i.e., DT with EEGNet. When each term is removed from the MOPC, the average accuracy drops from 81.23~\% to 80.32~\%, 79.67~\%, and 79.58~\%, respectively, with corresponding declines in kappa, which confirms the necessity of all three components. A similar trend is observed on OpenBMI, where omitting any term reduces accuracy by approximately 0.8~\%--1.4~\%, further indicating that perturbation-consistency regularization, joint entropy minimization, and confidence-calibrated reweighting each play a critical role in maximizing adaptation performance.

We further evaluated the impact of DPA and the dynamic memory bank, as shown in Table~\ref{tab:ablation_components}. Disabling DPA lowers both accuracy and kappa while removing the memory bank leads to an even larger decline in these metrics. The pronounced degradation when either component is omitted confirms that real-time alignment and memory-augmented context retention are essential for preventing catastrophic forgetting and ensuring robust online adaptation.

\begin{table}[!t]
  \centering
  \caption{Ablation study of the hybrid loss on BCI IV IIb and OpenBMI (\%)}
  \label{tab:ablation_results}
  \small
  \renewcommand{\arraystretch}{1.12}
  \setlength{\tabcolsep}{8pt}
  \begin{tabular}{c l c c}
    \hline\hline
    Dataset & Model & Acc $\pm$ std & Kappa $\pm$ std \\
    \hline
    \multirow{5}{*}{BCI IV IIb}
      & baseline                 & 76.28 $\pm$ 0.42 & 52.55 $\pm$ 0.83 \\
      & w/o $\mathcal{L}_{pcr}$  & 80.32 $\pm$ 0.24 & 60.64 $\pm$ 0.47 \\
      & w/o $\mathcal{L}_{jem}$  & 79.67 $\pm$ 0.15 & 59.33 $\pm$ 0.29 \\
      & w/o $\mathcal{L}_{ccr}$  & 79.58 $\pm$ 0.15 & 59.16 $\pm$ 0.31 \\
      & Whole                    & \textbf{81.23 $\pm$ 0.11} & \textbf{61.99 $\pm$ 0.23} \\
    \hline
    \multirow{5}{*}{OpenBMI}
      & baseline                 & 73.57 $\pm$ 0.34 & 50.07 $\pm$ 0.67 \\
      & w/o $\mathcal{L}_{pcr}$  & 77.56 $\pm$ 0.12 & 55.13 $\pm$ 0.25 \\
      & w/o $\mathcal{L}_{jem}$  & 77.75 $\pm$ 0.23 & 55.51 $\pm$ 0.46 \\
      & w/o $\mathcal{L}_{ccr}$  & 77.16 $\pm$ 0.29 & 54.31 $\pm$ 0.57 \\
      & Whole                    & \textbf{78.94 $\pm$ 0.14} & \textbf{57.88 $\pm$ 0.28} \\
    \hline\hline
  \end{tabular}
  \begin{flushleft}
    \footnotesize The best accuracies are marked in bold.
  \end{flushleft}
\end{table}

\begin{table}[!t]
  \centering
  \caption{Ablation of DPA and memory bank (\%)}
  \label{tab:ablation_components}
  \small
  \renewcommand{\arraystretch}{1.12}
  \setlength{\tabcolsep}{8pt}
  \begin{tabular}{c l c c}
    \hline\hline
    Dataset & Model & Acc $\pm$ std & Kappa $\pm$ std \\
    \hline
    \multirow{3}{*}{BCI IV IIb}
      & w/o DPA   & 80.03 $\pm$ 0.44 & 60.07 $\pm$ 0.87 \\
      & w/o Bank  & 77.65 $\pm$ 0.37 & 55.29 $\pm$ 0.75 \\
      & Whole     & \textbf{81.23 $\pm$ 0.11} & \textbf{61.99 $\pm$ 0.23} \\
    \hline
    \multirow{3}{*}{OpenBMI}
      & w/o DPA   & 77.86 $\pm$ 0.35 & 55.72 $\pm$ 0.69 \\
      & w/o Bank  & 76.69 $\pm$ 0.19 & 53.37 $\pm$ 0.38 \\
      & Whole     & \textbf{78.94 $\pm$ 0.14} & \textbf{57.88 $\pm$ 0.28} \\
    \hline\hline
  \end{tabular}
  \begin{flushleft}
    \footnotesize The best accuracies are marked in bold.
  \end{flushleft}
\end{table}

\subsubsection{Impact of Memory Bank Size}
\begin{figure}
	\centering
	\includegraphics[width=.45\textwidth]{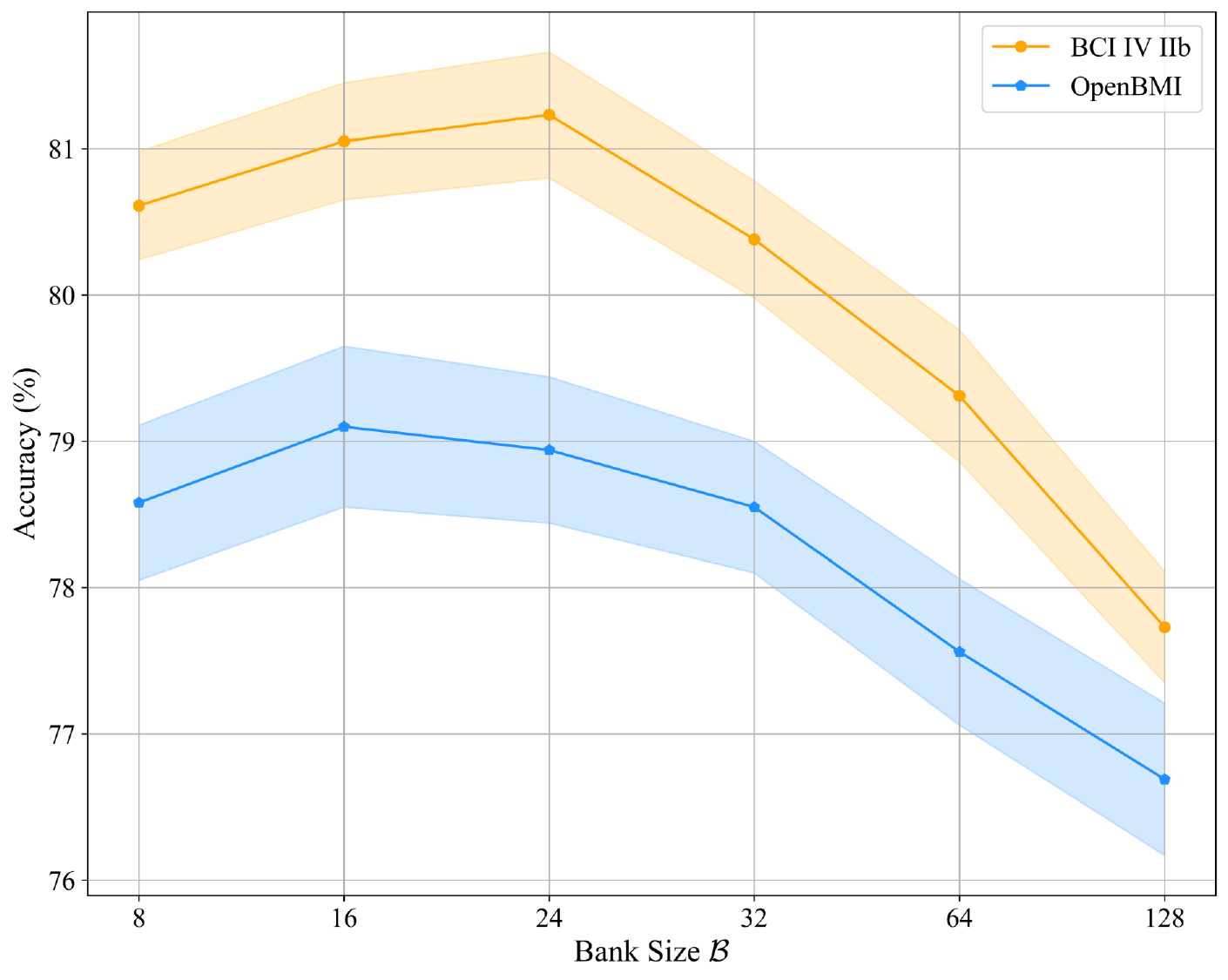}
	\caption{Effect of memory bank size on cross-subject accuracy.}
	\label{FIG:bank}
\end{figure}

We evaluated memory bank sizes \(\mathcal{B}\) ranging from 8 to 128 on BCI IV IIb and OpenBMI datasets, as presented in Fig. \ref{FIG:bank}. On BCI IV IIb, accuracy increased from 80.61~\% at size 8 to a peak of 81.23~\% at size 24, then declined. On OpenBMI, accuracy rose from 78.58~\% at size 8 to 79.10~\% at size 16, plateaued at 78.94~\% for size 24, and decreased thereafter. The shaded regions denote the standard deviation across five random seeds, reflecting run-to-run variability.

These trends highlight a tradeoff: a small bank provides insufficient context for stable adaptation, whereas an overly large bank retains outdated samples that hinder responsiveness to recent domain shifts. Based on these findings, we selected a memory bank size of 24 for all experiments, as it offers near-optimal accuracy on both datasets while keeping memory and computational overhead manageable.

\begin{table}[!t]
  \centering
  \caption{Performance and Computational Cost across Different Backbones}
  \label{tab:backbone_generalization_cost}
  \small
  \renewcommand{\arraystretch}{1.12}
  \setlength{\tabcolsep}{8pt}
  \begin{tabular}{l c c c c}
    \hline\hline
    Backbone & Baseline & MOPC & Parameters & FLOPs \\
    \hline
    EEGNet     & 76.28 & 81.23          & 2,060    & 2.87M \\
    ConvNet    & 73.12 & 76.96          & 10,802   & 7.66M \\
    LMDA       & 75.52 & 79.32          & 3,094    & 6.71M \\
    Conformer  & 74.70 & 78.60          & 839,797  & 20.15M \\
    \hline\hline
  \end{tabular}
\end{table}

\subsubsection{Model-Agnostic Generalization}
To verify the generality of our approach, we compare MOPC with the source-only setting and observe consistent improvements across all four backbones. The results are shown in Table~\ref{tab:backbone_generalization_cost}. Both the convolutional models (EEGNet, ConvNet) and the attention-based models (LMDA, Conformer) achieve clear gains. The original ranking is largely preserved, while the gaps narrow with MOPC, indicating that the gains arise from mitigating distribution shift rather than model-specific tuning. These benefits span convolutional and attention-based designs, showing that the framework transfers effectively to the target domain. Overall, MOPC is model-agnostic and plug-and-play, demonstrating the universality of our OTTA approach and its ability to enhance cross-subject generalization and improve the domain adaptability of classic MI classification frameworks.


\section{Interpretable Analysis and Discussion}
\label{sec:discussion}
\begin{figure}
	\centering
	\includegraphics[width=.45\textwidth]{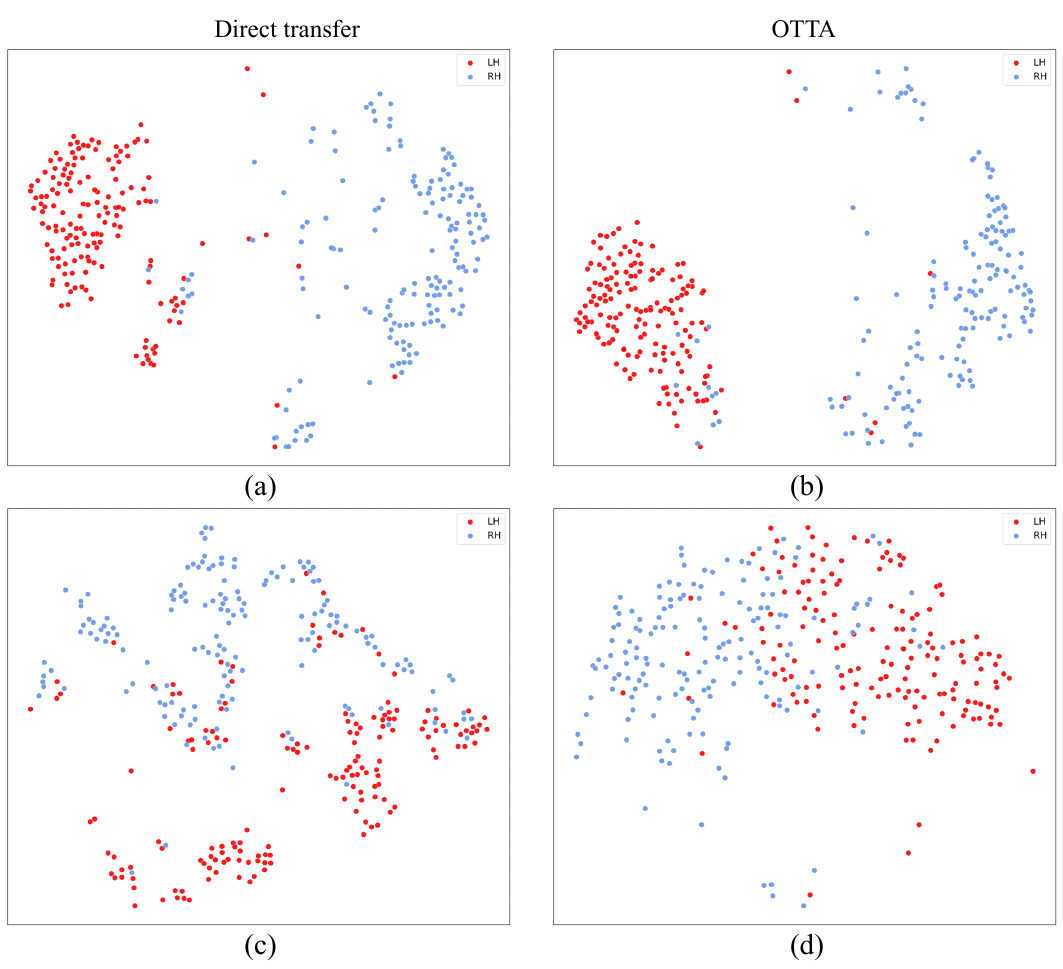}
	\caption{t-SNE embeddings of BCI IV IIb features for (a) and (b) subjects B04 and (c) and (d) B06. (a) and (c) DT, while (b) and (d) effect of OTTA.}
	\label{fig:tsne}
\end{figure}
\subsection{Feature Visualization} 
To evaluate feature space adaptation, we applied t-SNE to BCI IV IIb embeddings (subjects B04 and B06), as shown in Fig. \ref{fig:tsne}. Under DT [Fig.~\ref{fig:tsne}(a) and (c)], LH and RH imagery samples remain heavily intermingled due to severe cross-subject domain shifts. On the contrary, by dynamically updating the network's BN statistics via our unsupervised hybrid loss, MOPC effectively mitigates this variability. Consequently, the adapted embeddings [Fig.~\ref{fig:tsne}(b) and (d)] exhibit clearly separated LH and RH clusters. This sharpened decision boundary confirms that our calibration-free approach produces highly discriminative representations on unseen data, significantly enhancing online classification robustness for real-world BCI applications.

\subsection{Computational Efficiency and Real-Time Suitability}
To validate the feasibility for closed-loop instrumentation, we quantified the runtime latency on a standard workstation using an Intel Xeon Silver 4210R CPU. Quantitative analysis reveals that the total processing pipeline incurs a latency of approximately 40--60 ms per trial on a CPU. This pipeline includes preprocessing, inference, and the proposed online adaptation. This latency is low enough for responsive closed-loop feedback, for which keeping processing delay and timing jitter small is a recognized design concern in real-time BCI~\cite{wilson2010procedure}.

Table~\ref{tab:complexity} illustrates the comparison of the test-time cost of each adaptation method. MOPC and TENT are the only methods that restrict gradient updates to the BN affine parameters, whereas CoTTA, T-Time, and DPLOT update the full network, and SHOT adapts the entire encoder offline over the whole target set. MOPC also keeps adaptation independent of the class count, and its only additional state is a bounded FIFO memory of \mbox{$\mathcal{B}=24$} trials and an \mbox{$N=5$} ensemble. Although TENT shares this lightweight BN-only update, MOPC attains substantially higher cross-subject accuracy at the same update cost (Tables~\ref{tab:bci2b}~and~\ref{tab:openbmi}). The complete pipeline runs at 40--60 ms per trial, as reported above.

\begin{table}[!t]
  \centering
  \caption{Test-time computational profile of the adaptation methods, derived from each method's design (backbone-independent)}
  \label{tab:complexity}
  \footnotesize
  \renewcommand{\arraystretch}{1.25}
  \begin{tabular}{@{}l >{\raggedright\arraybackslash}p{2.2cm} >{\raggedright\arraybackslash}p{1.4cm} >{\raggedright\arraybackslash}p{1.0cm}@{}}
    \hline\hline
    Method & Params updated (test) & Backprop / update & Cost vs.\ \#cls. \\
    \hline
    Direct transfer & none & none & $\mathcal{O}(1)$ \\
    SHOT\textsuperscript{$\dagger$} & feature encoder & multi-epoch & $\mathcal{O}(K)$ \\
    TENT & BN affine only & 1 & $\mathcal{O}(1)$ \\
    CoTTA\textsuperscript{$\ddagger$} & full network & 1 & $\mathcal{O}(1)$ \\
    T-Time & full net, $\times M$ & $M$ & $\mathcal{O}(K)$ \\
    DPLOT & full network & 2 & $\mathcal{O}(1)$ \\
    MOPC (ours) & BN affine, 1 of $N{=}5$ & 1 & $\mathcal{O}(1)$ \\
    \hline\hline
  \end{tabular}
  \begin{flushleft}\footnotesize
  $K$: number of classes; $M$/$N$: ensemble size; $\mathcal{B}$: memory-bank size. All methods adapt online per trial except SHOT (\textsuperscript{$\dagger$}), which adapts offline over the whole target set for multiple epochs. \textsuperscript{$\ddagger$}CoTTA also runs up to 32 augmentation-averaged teacher passes per trial when teacher confidence is low, so its $\mathcal{O}(1)$-in-classes label does not imply a low per-trial cost. Extra state: SHOT keeps information-maximization statistics; CoTTA and DPLOT keep an EMA teacher (CoTTA also a source snapshot, DPLOT prototypes); T-Time keeps $M$ model copies $+$ EA state; MOPC keeps a $\mathcal{B}{=}24$ FIFO bank $+$ $N{=}5$ ensemble; TENT and direct transfer keep none. Backprop depth is the parameter subset receiving a gradient step per adapted trial; measured end-to-end MOPC latency is 40--60~ms/trial (this section).
  \end{flushleft}
\end{table}

Table~\ref{tab:backbone_generalization_cost} details the performance and overhead across various backbones. MOPC consistently improves decoding accuracy while keeping its added cost bounded rather than negligible: adaptation is confined to BN-affine updates over a fixed-size memory bank and a single member of an \mbox{$N=5$} ensemble per trial, so the parameter and memory overhead stays small enough for real-time use.

\subsection{Extension to Multiclass MI}
\label{subsec:multiclass}
MOPC extends from binary to multiclass MI without any architectural modification. Because each trial updates only the BN affine parameters of a single ensemble member, the number of trainable parameters is independent of the class count, so the adaptation step remains \mbox{$\mathcal{O}(1)$} in the number of classes (Table~\ref{tab:complexity}); moving to \mbox{$K$} classes changes only the output dimension of the classifier head, while the memory bank, the dynamic EA, and all three loss terms operate unchanged. This contrasts with methods whose test-time cost grows with the class count, such as the per-class information-maximization objectives of SHOT and T-Time. The three losses are themselves defined over the predictive distribution \mbox{$\boldsymbol{p}\in\mathbb{R}^{Y}$} and are agnostic to the number of classes: the perturbation-consistency and joint-entropy terms reduce to a KL divergence and a Shannon entropy over the \mbox{$Y$}-way softmax, and the confidence-calibrated reweighting assigns an effective-number weight per class, so it naturally accommodates additional classes and any greater imbalance they introduce.

To complement this structural argument, we ran a preliminary empirical check on the four-class BCI Competition IV 2a dataset under a LOSO protocol, applying MOPC with no architectural change: only the classifier head was widened to four outputs, while the ensemble (\mbox{$N=5$}), the memory bank, the dynamic EA, and all three loss terms were reused with the binary hyperparameters (\mbox{$\beta=0.999$}, \mbox{$\gamma=0.5$}, \mbox{$W=200$}, BN-affine-only updates, equal 1:1:1 weighting; 1000-epoch source models using the same EEGNet backbone as the binary experiments). Table~\ref{tab:bci2a_4class} summarizes the outcome: relative to a frozen source ensemble with static source-reference alignment, the full MOPC pipeline improves four-class accuracy for all nine subjects.

\begin{table}[t]
\centering
\caption{Preliminary four-class decomposition on BCI IV 2a (single-seed, LOSO). MOPC is compared only against its own source-only and alignment-only ablations; no competing test-time-adaptation method is evaluated here}
\label{tab:bci2a_4class}
\small
\begin{tabular}{@{}lccc@{}}
\hline
Condition & Acc.\ (\%) & $\kappa\ (\%)$ & $\Delta$ vs.\ source \\
\hline
Source-only (static EA)     & 39.97 & 19.96 & --- \\
\quad + dynamic EA (no BN)  & 52.56 & 36.75 & $+12.59$ \\
MOPC (dynamic EA + BN)      & 55.05 & 39.73 & $+15.08$ \\
\hline
\end{tabular}
\end{table}

To keep this gain interpretable, we decompose it with an intermediate condition, namely the same frozen ensemble with the online, memory-bank-derived target-reference alignment enabled but the BN adaptation switched off (Table~\ref{tab:bci2a_4class}, middle row). Dynamic alignment accounts for \mbox{$+12.59$} percentage points of the improvement (all nine subjects), while the online adaptation contributes the remaining \mbox{$+2.49$} points, improving six of the nine subjects and slightly regressing the other three (including the near-chance subject A02, whose source model sits close to the \mbox{$25\%$} level). The adaptation contribution is thus real but modest, and smaller than in the binary setting: we attribute this to the difficulty of four-class decoding, where pseudolabel-driven updates on near-chance subjects are fragile or mildly counterproductive, and the benefit concentrates on middifficulty subjects. The validation is preliminary and compares MOPC only against its own source-only and alignment-only ablations, so it establishes that the framework operates unchanged and improves over source transfer at four classes, but makes no claim of superiority over competing test-time-adaptation methods; a full multiclass benchmark against such methods, with multiseed statistics, remains future work.

\section{Conclusion}
\label{sec:conclusion}
In this study, we presented MOPC, a memory-augmented OTTA framework explicitly designed for robust EEG instrumentation in nonstationary measurement environments. By integrating perturbation consistency regularization, joint entropy minimization, and confidence-calibrated reweighting, MOPC enables efficient, source-free, single-pass cross-subject decoding. Extensive experiments on two public MI-BCI datasets show that MOPC consistently surpasses DT, DA, and state-of-the-art TTA baselines while maintaining highly stable cross-subject measurement reliability. Ablation studies confirm that each loss term and architectural component is integral to overall performance, contributing to robust and stable online adaptation. Furthermore, model updates incur fewer than 60 ms of latency on a standard CPU, satisfying real-time BCI constraints. However, current limitations include empirical validation centered on binary classification (with a preliminary four-class result confirming that the framework operates unchanged and improves over source transfer), while the design extends to multiclass decoding at \mbox{$\mathcal{O}(1)$} adaptation cost and relies on initial pseudolabel quality for effective reweighting. Future work will investigate adaptive augmentation strategies, multimodal physiological fusion, and comprehensive multiclass benchmarking, spanning four-class MI and continuous control, to further enhance robustness and practical applicability in real-world BCI systems.

\bibliographystyle{IEEEtranN}
\bibliography{refs}

\begin{IEEEbiography}[{\includegraphics[width=1in,height=1.25in,clip,keepaspectratio]{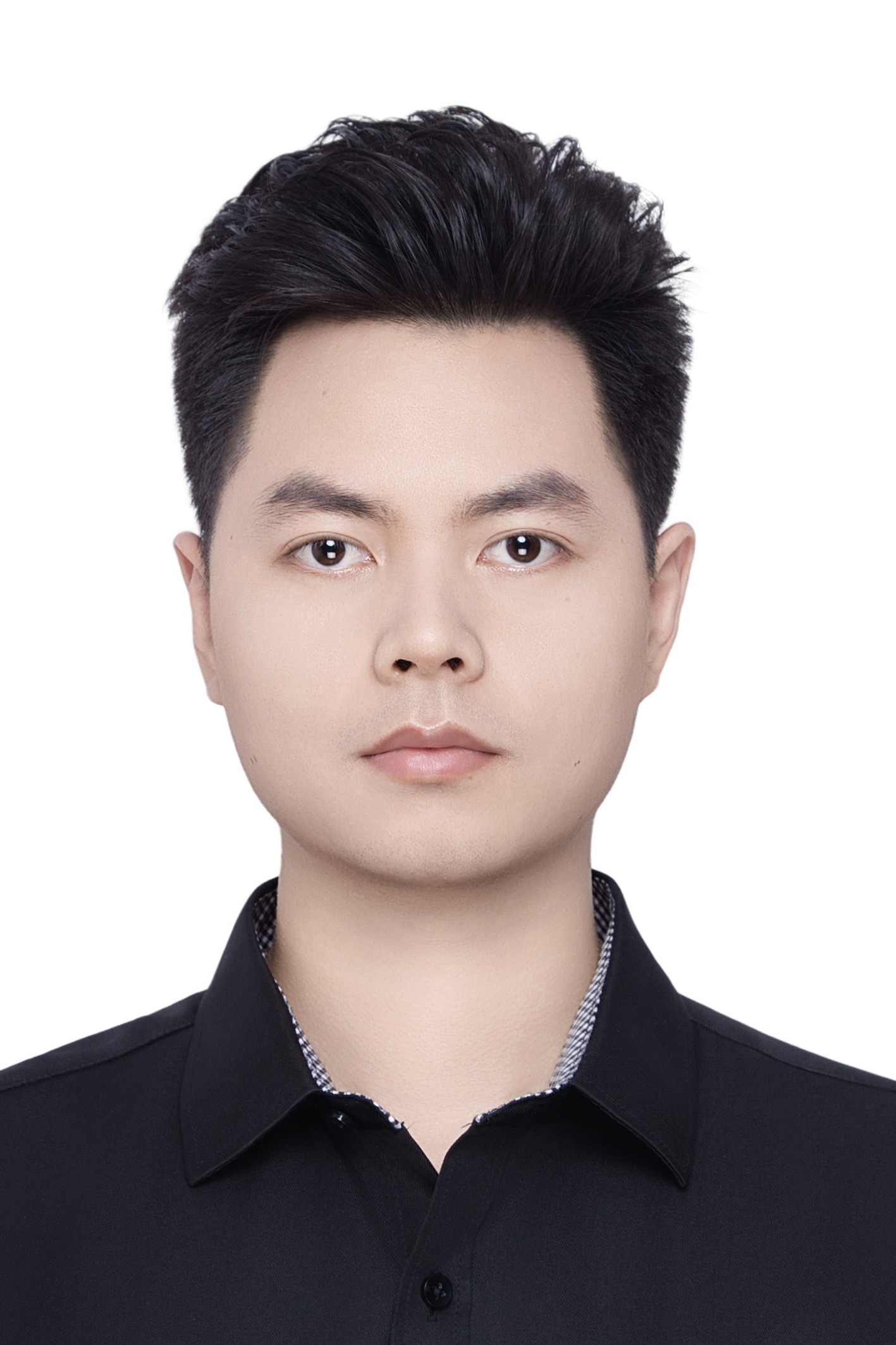}}]{Kunkun Zhang}
received the B.S. degree in mathematics from Nanyang Institute of Technology, Nanyang, China, in 2021. He is currently pursuing the Ph.D. degree in computer science with Zhejiang University of Technology, Hangzhou, China.

He is currently a Visiting Scholar with Weill Cornell Medicine, New York, NY, USA. His research interests include brain–computer interfaces, EEG decoding, and the measurement of cortical dynamics with electrophysiological and optical imaging techniques, with an emphasis on distribution shift and cross-session generalization of neural decoders.
\end{IEEEbiography}

\begin{IEEEbiography}[{\includegraphics[width=1in,height=1.25in,clip,keepaspectratio]{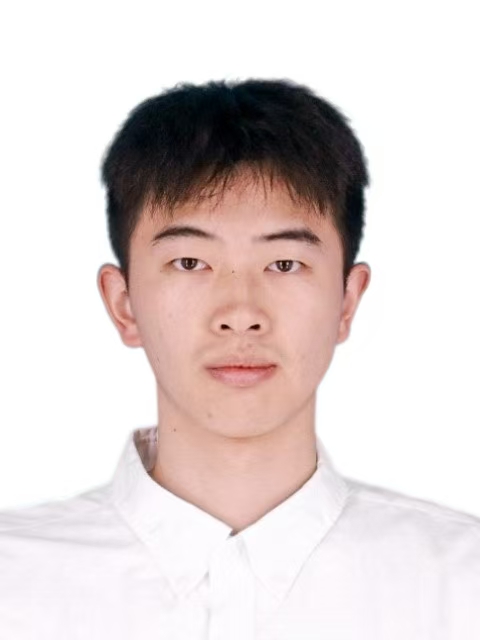}}]{Kehao Wu}
received the M.S. degree in computer technology from the School of Computer Science and Technology, Zhejiang University of Technology, Hangzhou, China, in 2026. 

His research interests include brain–computer interfaces, biomedical machine learning, and transfer learning, with a focus on online test-time adaptation for cross-subject EEG decoding in practical measurement systems.
\end{IEEEbiography}

\begin{IEEEbiography}[{\includegraphics[width=1in,height=1.25in,clip,keepaspectratio]{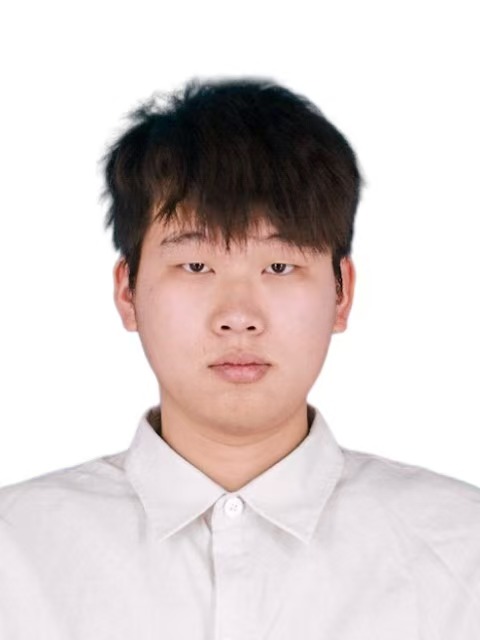}}]{Zhitao Chen}
received the M.S. degree in software engineering from the College of Computer Science and Technology, Zhejiang University of Technology, Hangzhou, China, in 2026. 

His research interests include brain–computer interfaces, motor imagery EEG decoding, and statistical manifold learning, with a focus on geometric methods for the analysis of nonstationary neurophysiological signals.
\end{IEEEbiography}

\begin{IEEEbiography}[{\includegraphics[width=1in,height=1.25in,clip,keepaspectratio]{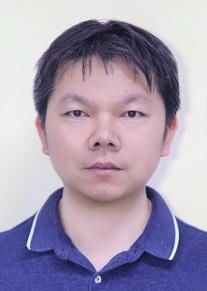}}]{Haigen Hu}
(Member, IEEE) received the Ph.D. degree in control theory and control engineering from Tongji University, Shanghai, China, in 2013. 

He has come out from the Post-Doctoral station of the LITIS Laboratory, University of Rouen Normandy, Rouen, France, in 2019. He is currently a Professor with the College of Computer Science and Technology, Zhejiang University of Technology, Hangzhou, China. His current research interests include machine learning, computer vision, and medical image processing. 
\end{IEEEbiography}

\begin{IEEEbiography}[{\includegraphics[width=1in,height=1.25in,clip,keepaspectratio]{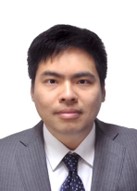}}]{Qianwei Zhou}
(Member, IEEE) received the Ph.D. degree in communication and information systems from Shanghai Institute of Microsystem and Information Technology, University of Chinese Academy of Sciences, Shanghai, China, in 2014.

In July 2014, he joined Zhejiang University of Technology, Hangzhou, China, where he is currently an Associate Professor with the College of Computer Science and Technology. He was a Visiting Scholar with the Imaging Research Division, Department of Radiology, University of Pittsburgh, Pittsburgh, PA, USA, from August 2018 to August 2019, and a Post-Doctoral Associate with the Department of Anesthesiology, Weill Cornell Medicine, New York, NY, USA, from September 2024 to September 2026, where he worked on simultaneous electrocorticography and cell-type-resolved calcium imaging of the rodent cortex. He has published scientific papers in international journals and conference proceedings, including Nature Communications, IEEE TRANSACTIONS ON INDUSTRIAL ELECTRONICS, IEEE SIGNAL PROCESSING LETTERS, and IEEE TRANSACTIONS ON MAGNETICS. He has been awarded grants from the National Natural Science Foundation of China (61802347 and 62271448). His research interests include robust EEG understanding, in particular cross-session and cross-subject generalization, robustness to electrode-side perturbations, such as channel reordering and partial electrode failure, and the use of concurrent optical and intracranial recordings as supervision for surface electrophysiology.
\end{IEEEbiography}

\end{document}